\documentclass[letterpaper, 10 pt, conference]{ieeeconf}  

\usepackage{cite}
\usepackage{bm}
\usepackage{makecell}
\usepackage{amsmath} 
\usepackage{amssymb}  
\usepackage[dvipsnames]{xcolor}
\usepackage{amsfonts}
\usepackage{amssymb}
\usepackage{amsmath}
\usepackage{graphics} 
\usepackage{epsfig} 
\usepackage{mathptmx} 
\usepackage{times} 
\usepackage{subfigure}

\usepackage{mathrsfs}
\usepackage{indentfirst}
\usepackage{multirow}
\usepackage{adjustbox}

\usepackage{algorithmic}
\usepackage{graphicx}
\usepackage{textcomp}
\usepackage{xcolor}
\usepackage{bbding}

\usepackage[colorlinks=true,linkcolor=blue,citecolor=green,urlcolor=blue,]{hyperref}

\usepackage{algorithm}
\usepackage{algorithmic}

\IEEEoverridecommandlockouts                              

\title{\LARGE \bf
Ground-Fusion: A Low-cost Ground SLAM System Robust \\ to Corner Cases
}

\author{Jie Yin, Ang Li, Wei Xi, Wenxian Yu, and Danping Zou*
\thanks{ All authors are with
Shanghai Jiao Tong University. $^*$ Corresponding Author: Danping Zou ({\tt\small dpzou@sjtu.edu.cn}). }%
\thanks{This work was supported by National Key R\&D Program (2022YFB3903802), NSFC(62073214), and Midea Group's 3D Vision Project.}
}

\begin{document}

\maketitle
\thispagestyle{empty}
\pagestyle{empty}

\begin{abstract}

We introduce Ground-Fusion, a low-cost sensor fusion simultaneous localization and mapping (SLAM) system for ground vehicles. Our system features efficient initialization, effective sensor anomaly detection and handling, real-time dense color mapping, and robust localization in diverse environments. We tightly integrate RGB-D images, inertial measurements, wheel odometer and GNSS signals within a factor graph to achieve accurate and reliable localization both indoors and outdoors. To ensure successful initialization, we propose an efficient strategy that comprises three different methods: stationary, visual, and dynamic, tailored to handle diverse cases. Furthermore, we develop mechanisms to detect sensor anomalies and degradation, handling them adeptly to maintain system accuracy. Our experimental results on both public and self-collected datasets demonstrate that Ground-Fusion outperforms existing low-cost SLAM systems in corner cases. We release the code and datasets at \href{https://github.com/SJTU-ViSYS/Ground-Fusion}{https://github.com/SJTU-ViSYS/Ground-Fusion}.

\end{abstract}

\section{INTRODUCTION}
Ground robots find extensive applications in logistics, catering, and industrial production. In many scenarios, it is critical to reliably navigate the ground robots within both indoor \cite{yin2023design} and outdoor environments\cite{cadena2016past}.
Simultaneous Localization and Mapping (SLAM) technology plays a key role in robot navigation. While LiDAR-based SLAM systems excel in many scenarios, their high costs are not suitable for low-cost applications where SLAM systems using affordable sensors such as Visual-Inertial Odometry (VIO) are preferred. However, VIO may exhibit reduced accuracy in specific motion modes that introduce unobserved degrees of freedom (DoFs), as discussed in \cite{Martinelli2013ClosedFormSO}. To tackle this challenge, VINS-RGBD \cite{shan2019rgbd} integrated RGBD images with inertial data to avoid scale issues and achieve better pose estimation. Similarly, VINS-on-Wheels \cite{wu2017vins} introduced a wheel odometer-assisted VIO system to maintain a consistent metric scale. \cite{hua2023m2c} \cite{yin2023robust} further loosely integrates GNSS signals into a visual-inertial-wheel-odometry.
Additionally, \cite{cao2022gvins} tightly combined GNSS raw measurements with the VIO system for drift-free global state estimation. \cite{yin2023sky} harnessed a sky-pointing camera for NLOS mitigation, thereby enhancing localization in urban canyons. 

Nevertheless, our previous investigations \cite{yin2021m2dgr} have revealed that the robustness of existing SLAM systems in challenging scenarios needs to be further improved. To address this concern, this paper focuses on two aspects: robust system initialization and corner case addressing. To ensure a successful initialization, we propose an effective strategy that includes three different methods: stationary, visual, and dynamic, which are designed to handle various situations. In addition, we discuss the possible sensor faults that might happen in corner cases \cite{yin2023ground} and handle them accordingly. 
We have conducted extensive experiments to evaluate our method. The results show the robustness of our method in different scenarios. We highlight the main contributions of this work as follows:

	\begin{figure}
		\centering
		\includegraphics[width=0.48\textwidth]{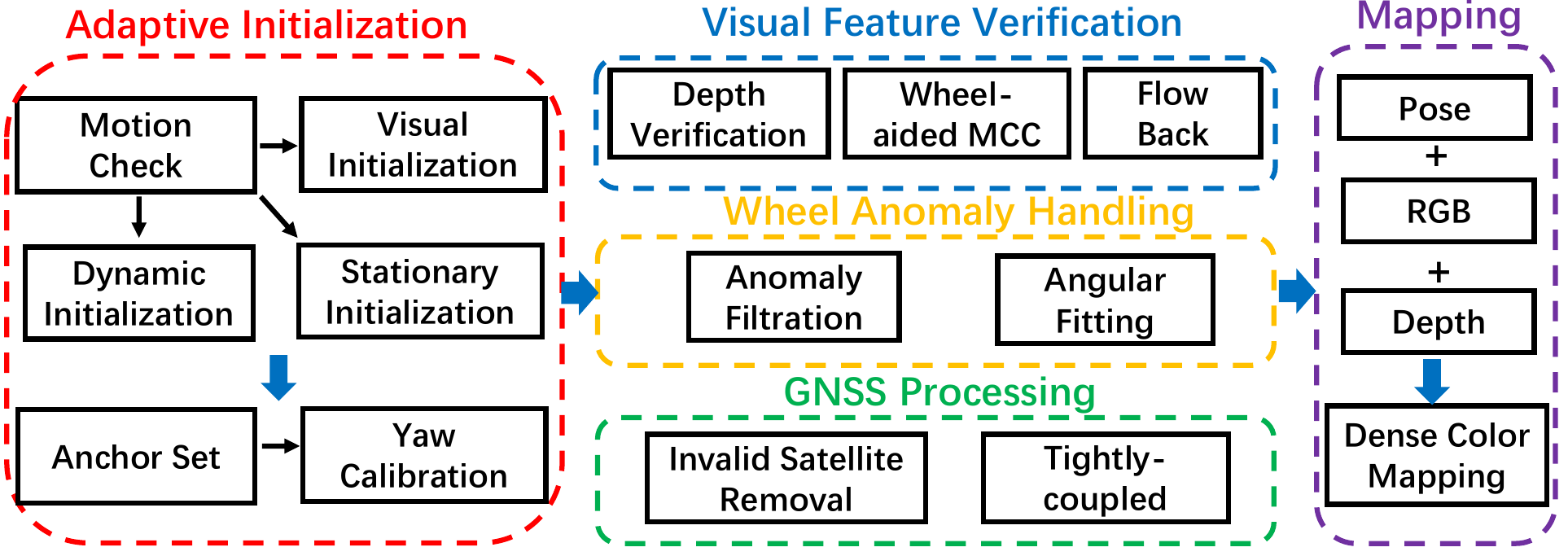}
  \vspace{-3mm}
		\caption{The system adopts an adaptive initialization strategy based on the robot's motion state. Potential sensor faults will be detected and handled accordingly. Real-time dense color mapping is supported to facilitate navigation tasks. }
		\label{pipeline}
	\end{figure}
\begin{itemize}

\item We implement a low-cost SLAM system that tightly couples GNSS-RGBD-IMU-Wheel sensors, which can work reliably both indoors and outdoors by fully exploiting each sensor to enable robust initialization in diverse cases.
\item  We propose effective strategies to detect and handle sensor faults that may arise in sensor fusion systems, including visual failures, wheel anomalies, and GNSS degradation, thereby greatly enhancing robustness.
\item  We present a SLAM dataset serving as a new benchmark for challenging corner cases.
\end{itemize}

\section{Related work}
\subsection{Initialization of a multi-sensor SLAM system}
Multi-sensor SLAM systems, particularly those that are tightly-coupled, are heavily reliant on high-quality initialization due to its profound impact on system robustness and accuracy. VINS-Mono \cite{qin2018vins} employs a hybrid initialization strategy that performs vision-only Structure from Motion (SfM) and a subsequent visual-IMU alignment. Then the scale and gravity direction will be further optimized. 
This method requires sufficient IMU excitation and visual parallax for successful initialization. However, the movement of a ground robot can easily experience unobservable DoFs, such as uniform motion, which makes the VIO systems unobservable at scale.
To mitigate this issue, \cite{shan2019rgbd} utilizes depth information as the source of the scale information for initialization. Similarly, \cite{liu2019visual} uses wheel odometer measurements to refine the gravity vector, utilizing visual SfM for scale estimation.
Moreover, \cite{gui2023zupt} proposes a Zero Velocity Update (ZUPT) aided initialization method with stationary and dynamic phase to achieve better robustness. However, the initialization of these systems is still not robust enough to severe sensor faults.
Building upon these insights, we introduce an adaptive initialization strategy encompassing three distinct approaches tailored to different scenarios in this work.


\subsection{Corner cases}
\textbf{Visual challenge:} 
We classify vision failures into three categories. The first type is insufficient-features problem, typically caused by lack of textures or inadequate light. This can lead to significant drift in visual systems \cite{qin2018vins, campos2021orb}. The second is characterized by no valid feature points due to significant occlusion or aggressive motion. This can result in unsuccessful system initialization or tracking failure as demonstrated in \cite{yin2023ground}. The third type refers to dynamic environments with numerous moving objects. The moving points on these objects can greatly degrade localization accuracy. Currently, there exist some effective semantic-based methods \cite{liu2022rgb} \cite{yu2018ds} that well tackle this challenge. In this study, we focus on cost-efficient geometric approaches that do not utilize GPUs. Specifically, we reject dynamic features through feature filtration and depth validation.

\textbf{Wheel odometer challenge:}
There are two types of wheel odometer challenges: inaccurate angular velocity and wheel anomaly. To tackle the former issue, \cite{quan2019tightly} integrated measurements from wheel encoders and gyroscopes to formulate a relative motion constraint.
For the latter challenge, they detect the wheel anomaly with visual information: If more than half of the previous features are considered as outliers during visual-wheel optimization, wheel slippage is detected. In response, they reset the current frame's initial state and eliminate current wheel measurements to mitigate the anomaly.
However, a significant decrease in feature points could also arise from environmental changes, rather than just wheel-related factors. Therefore, this strategy is likely to mistakenly eliminate reliable wheel odometer observations. Considering this limitation, \cite{peng2020pose} introduced three techniques for actively detecting abnormal chassis movements with the help of motion constrains and IMU measurements. We integrate insights from these efforts to enhance our system's capacity to address wheel odometer challenges.

\textbf{GNSS degradation:}
Typically, there are three kinds of GNSS challenges: low-speed movements, less than four satellites, and no GNSS signals.
Early work employed Receiver Autonomous Integrity Monitoring (RAIM) scheme \cite{hewitson2006gnss} to assess integrity performance levels of GNSS systems, detecting and mitigating errors within the GNSS receiver based on residuals. As a GNSS-Visual-IMU tightly-coupled system, GVINS \cite{cao2022gvins} proposes strategies to address GNSS-related issues in challenging scenarios. GVINS \cite{cao2022gvins} remains robust in less-than-four-satellite case due to its tighly-coupled framework. In GNSS-denied environments, GVINS degenerates into a convectional VIO. Building upon these insights, our system firstly filters out unreliable satellites with threshold-based methods. Furthermore, our system monitors low-speed states and ensures that GNSS-related factors are not optimized in these scenarios.

\textbf{IMU saturation:} He et al. \cite{he2023point} address IMU saturation issues for drones with highly aggressive motions, while this situation is rarely encountered by safety-critical ground robots. In our study, we consider short-term IMU observations within a sliding window to be reliable and free from faults.

\subsection{Ground robot datasets}
A high-quality dataset can stimulate progress in SLAM. While there have been a few datasets captured by ground robots \cite{sturm2012benchmark} \cite{leung2011utias} \cite{shi2020we}, many of them are outdated and not challenging enough to current advanced SLAM systems. 
Our previous work \cite{yin2021m2dgr} presents a challenging dataset M2DGR
\footnote{\href{https://github.com/SJTU-ViSYS/M2DGR}{https://github.com/SJTU-ViSYS/M2DGR}}, which has been met with great interest from SLAM researchers. Therefore, we further extend the M2DGR dataset by adding extra low-cost sensors, including a wheel encoder, a 2D LiDAR and a depth camera. This not only allows us to test our proposed system, but also launches a novel benchmark for future research on multi-sensor SLAM.

\section{Methodology}

Our Ground-Fusion system tightly integrates RGB images, depth information, inertial information, and wheel odometer measurements within the optimization framework.
All the sensory measurements are maintained in a sliding window for real-time performance. The system consists of adaptive initialization, a multi-sensor state estimator with corner case addressing, and a dense mapping module.
Before introducing each module, we clarify the notations and frames used in this paper: $(\cdot)^w$ denotes the world frame. $(\cdot)^b$ is the body (IMU) frame, and $(\cdot)^c$ is the camera frame. $\mathbf{q}_b^w$ and $\mathbf{p}_b^w$ represent the rotation and translation from the body frame to the world frame, respectively. $b_j$ and $c_j$ are the body frame and the camera frame when capturing the $j$th image. 

\subsection{Adaptive initialization}
Our initialization module consists of three alternative strategies for local odometry initialization: dynamic method, visual method and stationary one. The system determines whether there is sufficient motion excitation based on the GLRT (Generalized Likelihood Ratio Test) \cite{skog2010zero} method, which is formulated as:
\vspace{-2mm}

\begin{equation}
\small 
\mathbb{G}=\frac{1}{m} \sum_{t \in \psi}\left(\frac{1}{\sigma_a^2}\left\|\tilde{a}_t-g \frac{\bar{a}_t}{\left\|\bar{a}_t\right\|}\right\|^2+\frac{1}{\sigma_\omega^2}\left\|\tilde{\omega}_t\right\|^2\right) 
\vspace{-1mm}
\end{equation}
where $\left\{\tilde{{a}}_{t}, \tilde{\boldsymbol{\omega}}_{t}\right\}$ are the raw measurements of the IMU, $m$ is IMU measurement number within the sliding window, $\psi$ denotes the window range, and $\bar{a}$ is the average acceleration.



The GLRT value roughly divides the motion state into the following three categories, where
the threshold values for $\beta$ and $\gamma$ are determined by experimental approach:

\begin{equation}
\small 
\mathrm{State}\approx \left\{\begin{array}{cc}
\text { Stationary } & (\mathbb{G} < \beta)\\
\text { Slow Motion } & (\beta \leq \mathbb{G} \leq {\gamma} \text )\\
\text { Aggressive Motion} & (\mathbb{G} > \mathbb{\gamma})
\end{array}\right.
\vspace{-2mm}
\end{equation}
Next, we'll further validate the motion state and describe the corresponding initialization method for each kind of motion.


\textbf{Stationary:}
If the $\mathbb{G}$ is below the value $\beta$, we introduce wheel and visual observations to further ensure whether the system is static. We utilize wheel median integration method to predict the pose:
\vspace{-2mm}
\begin{equation}
\small 
\begin{aligned}& 
\mathbf{R}^o_{j+1}=\mathbf{R}_j^o \operatorname{Exp}\left(\overline{\bm{\omega}}^o_{j+1} \Delta t\right) \\
& \bm{p}_{{j+1}}^w=  \bm{p}_{{j+1}}^w+\overline{\bm{v}}^w_{j+1}\Delta t     
\end{aligned}
\end{equation}
\vspace{-1mm}
where $\Delta t$ is the time difference between odometer frames $o_j$ and $o_{j+1}$, and $\overline{\bm{\omega}}^o_{j+1} =\frac{1}{2}\left(\bm{\omega}^o_j+\bm{\tilde{\omega}}^o_{j+1}\right)$ and $\overline{\bm{v}}^w_{j+1} =\frac{1}{2}\left(\bm{v}^w_j+\bm{\tilde{v}}^w_{j+1}\right)$. Assuming there are $n$ odometer frames between consecutive images $c_k$ and $c_{k+1}$, the wheel preintegration pose's norm between them 
can be expressed as:
\vspace{-2mm}
\begin{equation}
\small 
\mathbb{W}=\left\|\bm{p}_{{j+n}}^w-\bm{p}_{{j}}^w\right\|^2
\end{equation}
\vspace{-1mm}
Additionaly, we can extract feature points from the latest frame match them with images in the sliding window.
The average visual parallax can then be formulated as:
\vspace{-1.5mm}
\begin{equation}
\small 
\mathbb{V}=\frac{1}{m} \sum_{i \in [0,m-1]}\left(\sum_{j \in [0,r-1]}\left\|\bm{p}_{i}^j-\bm{p}_{m-1}^j\right\|^2\right) 
\end{equation}
where $r$ is the matched feature point number between $j$th image and the latest image.

In the initialization phase, if at least two of the stationary criteria $\left\{\mathbb{G} < \beta, \mathbb{W} < \eta, \mathbb{V} < \theta \right\}$ are met (all threshold values determined by experimental approach), we consider the vehicle is static. Otherwise, we treat the vehicle as in motion and use the methods in the next paragraph for initialization. 
In the confirmed stationary case, we establish the first camera frame as the local world frame and align the $z$-axis with the gravity direction. Subsequently, all other poses within the sliding window are aligned with the first pose, while the velocity is set to zero. The system state $\bm{p}, \bm{v}, \bm{q}$ will be set to a constant block during optimization. The stationary detection and ZUPT is not only applicable to the initialization phase, but also used throughout the optimization process.

\textbf{Slow motion:}
In the slow motion case, 
the camera pose $\left(\bm{p}_{c}^w, \bm{q}_{c_k}^w\right)$ between the two frames could be computed by solving a PnP (Perspective-n-Point) problem. Since the RGB-D camera can directly measure the depth information, the IMU pose can be calculated without scale parameter by:
\vspace{-1.5mm}
\begin{equation}
\small 
\begin{aligned}
& {q}_{b_k}^w={q}_{c_k}^w \otimes\left({q}_{c_k}^{b_k}\right)^{-1} \\
& {p}_{b_k}^w={p}_{c_k}^w-{R}_{b_k}^w {p}_{c_k}^{b_k}
\end{aligned}
\vspace{-1.5mm}
\end{equation}
where the extrinsic $\left(\bm{p}_c^b, \bm{q}_c^b\right)$ is provided offline.

Combining above states with the IMU pre-integration term $\gamma$, we can calibrate the gyroscope bias by minimizing following least-square function:
\begin{equation}
\small 
\begin{array}{r}
\min _{\delta b_w} \sum_{k \in {B}}\left\|{q}_{b_{k+1}}^{c_0}{ }^{-1} \otimes{q}_{b_k}^{c_0} \otimes \boldsymbol{\gamma}_{b_{k+1}}^{b_k}\right\|^2 \\
\gamma_{b_{k+1}}^{b_k} \approx \hat{\gamma}_{b_{k+1}}^{b_k} \otimes\left[\begin{array}{c}
1 \\
\frac{1}{2} {J}_{b_w}^\gamma \delta {b}_w
\end{array}\right]
\end{array}
\end{equation}
where ${B}$ represents all frame indexes in the window
Since the scale is known, later parameters initialization only contain velocities and gravity vector. 
\begin{equation}
\small 
\boldsymbol{X}_I=\left[\bm{v}_{b_0}^{b_0}, \bm{v}_{b_1}^{b_1}, \cdots, \bm{v}_{b_n}^{b_n}, \boldsymbol{g}^{C_0}\right]
\end{equation}

Considering two consecutive IMU frames ${b_k}$ and $b_{k+1}$, we have following equations:


\vspace{-2mm}
\begin{equation}
\small 
\begin{aligned}
\boldsymbol{\alpha}_{b_{k+1}}^{b_k} & =\mathbf{R}_{c_0}^{b_k}\left(\overline{\bm{p}}_{b_{k+1}}^{C_0}-\overline{\bm{p}}_{b_k}^{c_0} - \mathbf{R}_{b_k}^{c_0} \bm{v}_{b_k}^{b_k} \Delta t + \frac{1}{2} \bm{g}^{c_0} \Delta t^2\right) \\
\bm{\beta}_{b_{k+1}}^{b_k} & =\mathbf{R}_{c_0}^{b_k}\left(\mathbf{R}_{b_{k+1}}^{c_0} \bm{v}_{b_{k+1}}^{b_{k+1}} - \mathbf{R}_{b_k}^{c_0} \bm{v}_{b_k}^{b_k} + \bm{g}^{c_0} \Delta t\right) 
\end{aligned}
\end{equation}
\vspace{-2mm}

Combining equation (6) and equation (9), we can solve the initial values of $\boldsymbol{X}_I$. Finally, the gravity vector obtained from the
previous linear initialization step is further refined.

\textbf{Aggressive motion:}
In highly aggressive motion, the visual features may be unstable due to motion blur or few overlap, causing the visual SfM-based initialization to be unreliable. By contrast,
the wheel odometer measurement makes the velocity and scale observable, the pose can be calculated by wheel integration. Consequently, we opt not to employ visual SfM for pose estimation as described in Equation $(6)$, but instead, we employ a wheel-aided initialization method. The camera pose in equation $(9)$ could be replaced by wheel odometer pose to solve the gyroscope bias. To establish a consistent reference frame, we define the world frame using the first wheel frame, aligning its z-axis with that of the wheel frame. In comparison to the approach used in \cite{liu2019visual}, which exclusively employs the wheel odometer for scale refinement, our method eliminates the redundant SfM component, fully harnessing the wheel odometer for a more efficient initialization process. It's worth noting that while this initialization method does not rely on visual information, once successfully initialized, the visual factor can still be integrated into the tightly coupled optimization process when the system identifies effective feature points. 
After successful local initialization by any of above three methods, we perform a three-step global initialization, which are adapted from \cite{cao2022gvins}.

\subsection{Multi-sensor state estimator with corner case handling}
We formulate the state estimation as a maximizing a posteriori (MAP) problem. We follow the factor graph framework of \cite{wu2017vins} which maintains a sliding window, and further extend to a GNSS-RGBD-IMU-Wheel fusion system. The calculation of residuals and Jacobi can refer to the previous literature \cite{cao2022gvins}\cite{shan2019rgbd}\cite{wu2017vins}. Next, we mainly introduce how our system processes sensor measurements to become more robust to corner cases.

\textbf{Wheel anomalies:}
The wheel odometer measurement can be formulated as $\left\{\tilde{\boldsymbol{v}}^o_t=\boldsymbol{v}^o_t+\boldsymbol{n}^v_t, \tilde{\boldsymbol{\omega}}^o_t=\boldsymbol{\omega}^o_t+\boldsymbol{n}^{\boldsymbol{\omega}}_t\right\}$.
Here $\left\{\tilde{{v}}_{t}, \tilde{\boldsymbol{\omega}}_{t}\right\}$ are the raw measurements of the wheel odometer, $\boldsymbol{v}^o=\left[\begin{array}{lll}v_x^o & v_y^o & 0\end{array}\right]^T, \boldsymbol{\omega}^o=\left[\begin{array}{lll}0 & 0 & \omega_z^o\end{array}\right]^T$ denote the velocity and angular velocity in the wheel odometer frame

The error of wheel odometer mainly comes from inaccurate angular velocity estimation and sudden chassis anomaly, such as wheel slippage and collision. 
Since the IMU's angular velocity measurement is more reliable and has a higher frame rate than the wheel odometer, we replace the original wheel odometer measurement with the IMU angular velocity using a linear fitting method:

\begin{equation}
\small 
\begin{aligned}
 {\boldsymbol{\omega}}^o_z={\mathbf{R}^o_b   (\boldsymbol{\omega}^b_m + \frac{\boldsymbol{\omega}^b_n-\boldsymbol{\omega}^b_m}{t_n-t_m}(t-t_m)-\bm{b}_\omega )}_z
\end{aligned}
\end{equation}
where $t_m$ and $t_n$ are two closest IMU timestamps to current wheel measurement.

To detect wheel anomalies, we compare the pre-integration of both the IMU and the wheel odometer measurements between the current frame and the second latest frame. If the difference of their resulting poses' norm surpasses the threshold $\epsilon=0.015$, we see it as a wheel abnormality. In this case, we refrain from incorporating the current wheel odometer observations into subsequent optimization process.

\textbf{Vision anomalies:}
Our system employs the KLT sparse optical flow algorithm \cite{lucas1981iterative} for tracking feature points as adapted from \cite{qin2018vins}. Three visual challenges include no-valid-feature problem during initialization, insufficient-feature issue during the localization and dynamic environments
The first one has been solved in Sec III (a), while the second one can be mitigated by tightly-coupled wheel odometer and IMU data. To address dynamic objects, 
We further introduce two strategies: feature filtration and depth validation.

For feature filtration, we firstly adopt the flow back method by reversing the order of the two frames for optical flow backtracking. Only the feature points that are successfully tracked during both iterations and exhibit an adjacent distance below a specified threshold are retained for further process. 
Moreover, we introduce a wheel-assisted moving consistency check (MCC) approach. Our system utilizes the wheel-preintegration pose and previous optimized poses. 
For a feature observed for the first time in the $i$-th image and subsequently observed in other $m$ images within the sliding windows, we define the average reprojection residual $r_k$ of the feature observations as:

\vspace{-1.5mm}
\begin{equation}
\small 
\bm{r}_k=\frac{1}{m} \sum_{j \neq i}\left\|\bm{u}_k^{c_i}-\pi\left(\mathbf{T}_b^c \mathbf{T}_w^{b_i} \mathbf{T}_{b_j}^w \mathbf{T}_c^b \mathbf{P}_k^{c_j}\right)\right\|
\end{equation}
\vspace{-1mm}
here $\mathbf{u}_k^{c_i}$ represents the observation of the $k$-th feature in the $i$-th frame, and $\mathbf{P}_k^{c_j}$ is the 3D location of the $k$-th feature in the $j$-th frame. The function $\pi$ denotes the camera projection model. When the value of $r_k$ exceeds a preset threshold, the $k$-th feature is considered dynamic and will be removed from further process. This approach offers the advantage of preemptively eliminating potential error-tracking features in current image frame before the optimization phase.

For the depth validation, we initiate by associating the depth measurements acquired from the depth camera with each pixel representing a feature point. In cases where the depth measurement surpasses the effective range of the depth camera, the pixel is temporarily left empty. Subsequently, we employ the triangulation method on the RGB image to calculate the depth of the feature points, thus filling in all the pixel depths. Additionally, for those feature points where the disparity between the depth measured by the depth camera and the depth computed through triangulation is below a predefined threshold, we record their indexes and fix their depth values to a constant value during the optimization phase. 

\textbf{GNSS anomalies:}
Three GNSS challenge scenarios include too-few-satellites, no-satellite-signal, and low-speed movement. In the first two cases, \cite{cao2022gvins} has proven that with the help of the tightly-coupled GNSS-Visual-Inertial framework, limited reliable satellites can still be effective in improving the global state estimation, and the GVIO system will degrade into a VIO system when no GNSS signals observed. In this work, our system firstly filters out unreliable satellites with excessive pseudo-range and Doppler uncertainty, those with an insufficient number of tracking times, and those at a low elevation angle. In low-speed scenarios with a GNSS receiver velocity below the threshold $v_{ths} = 0.3 m/s$ (the noise level of the Doppler shift),
we do not involve GNSS factors in the optimization to prevent GNSS noise from corrupting the state estimation.

\section{Experiments}
\subsection{Benchmark tests}
\textbf{Localization performance:}
Openloris-Scene \cite{shi2020we} is a SLAM dataset collected by a ground robot with an RGBD camera, an IMU and a wheel odometer.
Ground-Fusion is tested against cutting-edge SLAM systems on three scenarios of Openloris-Scene \cite{shi2020we}, namely Office (7 seqs), home (5 seqs) and corridor (3 seqs). Table~\ref{openloris result} shows that Ground-Fusion performs well on these scenarios.

\textbf{Initialization performance:} We conduct initialization tests on Ground-Challenge dataset \cite{yin2023ground} with complex sequences in corner cases\footnote{\href{https://github.com/sjtuyinjie/Ground-Challenge}{https://github.com/sjtuyinjie/Ground-Challenge}}. 
To evaluate the efficiency of system initialization, we measure the time required for each system to complete the initialization process, which is defined as the difference between the timestamp of the first observation received by the system and that of the first output pose.
In terms of the quality of initialization, we evaluate the Absolute Trajectory Error (ATE) RMSE \cite{sturm2012benchmark} for each system, focusing on the initial 10 seconds of each sequence.
We select several challenging sequences from Ground-Challenge \cite{yin2023ground} for the initialization test. These sequences include $Office3$, which features changing light conditions; $Darkroom2$ recorded in a dark room; and $Wall2$ captured in front of a textureless wall. 
Additionally, $Motionblur3$ exhibits severe camera motion blur during aggressive movement, while $Occlusion4$ involves severe camera occlusion. And $Static1$ was recorded in a stationary state with wheel suspension.

We evaluate our method against baseline methods on these challenging sequences. Results in Table~\ref{ini tab} demonstrate that our method excels in both quality and efficiency of initialization.
Specifically, in Sequence $Office3$, our system adaptively chooses visual initialization method, requiring the least amount of time.
 In visual challenging scenarios with limited textures ($Darkroom2$ and $Wall2$) and visual failure($Motionblur3$ and $Occlusion4$), our system adopts the dynamic initialization method and initialize successfully with high efficiency, while some baseline systems fail due to inadequate feature points for visual SfM. In stationary case $Static1$,
our system initializes successfully with the stationary method, while
baseline methods fail due to insufficient motion excitation.

\begin{table}[h]
    \caption{Average ATE RMSE(m) of SLAM systems on challenging sequences of Openloris-Scene \cite{shi2020we} dataset}
    \label{openloris result}
    \centering
    \begin{adjustbox}{width=\columnwidth}
        \begin{tabular}{cccccccc}
            \hline
            & {ORBSLAM3\cite{campos2021orb}} & {DSO\cite{engel2017direct}} & {VINS-Mono\cite{qin2018vins}} & {InfiniT\cite{kahler2015very}} & {Elastic-F\cite{whelan2015elasticfusion}} & {Ours} \\
            \hline
            {office}    & 0.52 & 0.75 & 0.19 & 0.16 & \textbf{0.13} & 0.18 \\
           {home}      & 1.11 & 0.67 & 0.54 & 0.58 & 1.45 & \textbf{0.46} \\
           {corridor}  & 2.98 & 10.66 & 3.41 & 3.43 & 17.76 & \textbf{0.71} \\
            \hline
        \end{tabular}
    \end{adjustbox}
    \vspace{-3mm}
\end{table}
\vspace{-3mm}






		

	\begin{table}[h]
		\caption{  Initialization Time Cost (s) and ATE RMSE(m) of SLAM systems on sample sequences}
		\label{ini tab}
		\centering
		\begin{adjustbox}{width=0.9\columnwidth}
		\begin{tabular}{cccccc}
			\hline
			\makecell[c]{ Sequence } &\makecell[c]{VINS-Mono \cite{qin2018vins}} & \makecell[c]{VINS-RGBD \cite{shan2019rgbd}}  & \makecell[c]{VIW-Fusion }    & \makecell[c]{Ours}\\
			\hline

		\makecell[c]{Office3}  & \makecell[c]{2.09 / \textbf{0.02}}  & \makecell[c]{ 2.10 / \textbf{0.02}} & \makecell[c]{2.42 / \textbf{0.02}}  & \makecell[c]{\textbf{1.37} / \textbf{0.02}}  \\

			  \makecell[c]{Darkroom2}   & \makecell[c]{\textbf{1.30} / 2.06}  & \makecell[c]{ 2.52 / 0.04} & \makecell[c]{1.61 / 0.04}  & \makecell[c]{{1.36} / \textbf{0.03}} \\

			\makecell[c]{Wall2} & \makecell[c]{2.50 / 0.08}  & \makecell[c]{\textbf{1.27} / 0.07 } & \makecell[c]{2.00 / 0.05}  & \makecell[c]{{1.37} / \textbf{0.03}}\\

			\makecell[c]{Motionblur3} & \makecell[c]{fail}  & \makecell[c]{ fail} & \makecell[c]{1.30 / 0.11}  & \makecell[c]{\textbf{1.30} / \textbf{0.09}}\\
			
			\makecell[c]{Occlusion4}   & \makecell[c]{fail}  & \makecell[c]{ fail} & \makecell[c]{fail}  & \makecell[c]{\textbf{5.52} / \textbf{0.14}} \\
			
			\makecell[c]{Static1}  & \makecell[c]{fail}  & \makecell[c]{ fail} & \makecell[c]{fail}  & \makecell[c]{\textbf{1.42} / \textbf{0.00}} \\

			\hline

		\end{tabular}
		
		\end{adjustbox}
		\vspace{-3mm}
	\end{table}

	\begin{figure}
		\begin{center}
			\begin{tabular}{cc}

		
				\includegraphics[scale=0.24]{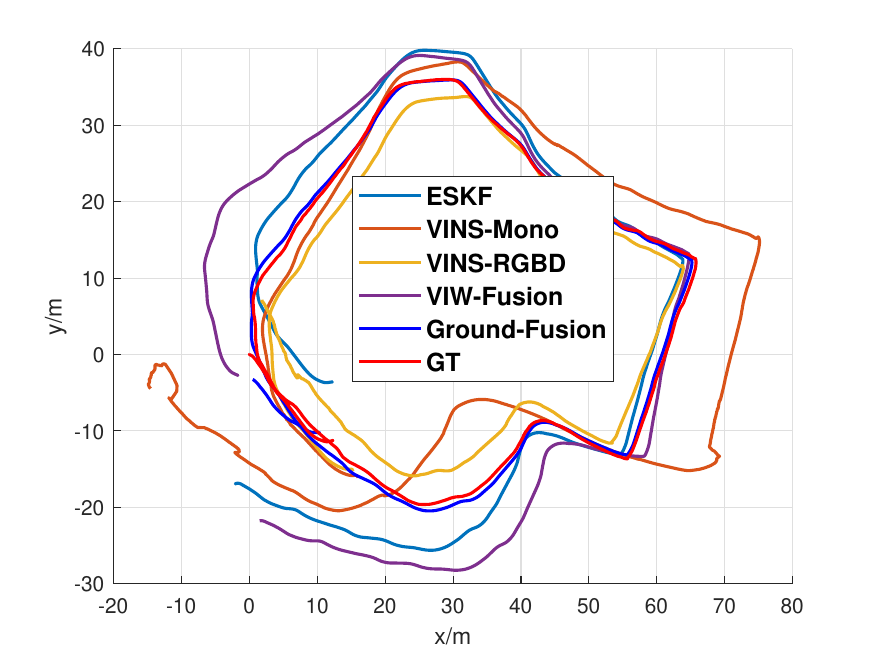}&
				\includegraphics[scale=0.24]{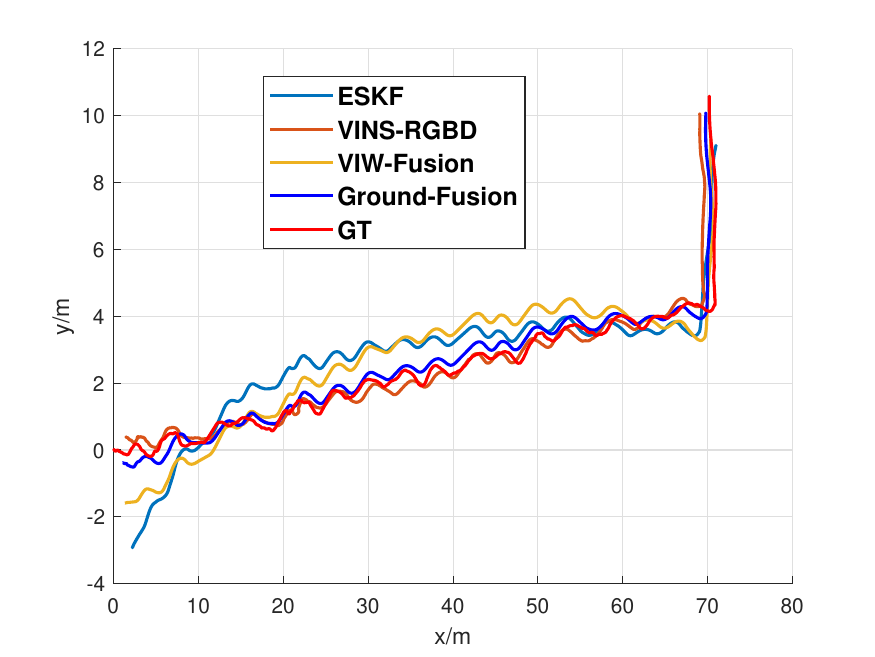}
				
				\\
				 (a).  Loop2 &(b). Corridor1
			
			\end{tabular}
			\vspace{-3mm}
		\end{center}
		\caption{Estimated and ground-truth (GT) trajectories on part of sample sequences are visualized on the x-y plane. }
		\vspace{-3.5mm}
		\label{plots}
	\end{figure}

	\begin{table}[h]
	
		\caption{ATE RMSE(m) of SLAM systems on sample sequences}
		\label{ate rmse tab}
		\begin{adjustbox}{width=\columnwidth}
		\centering
		\vspace{-1.5mm}
		\begin{tabular}{cccccccc}
			\hline
			\makecell[c]{ Sequence } & \makecell[c]{VINS-Mono \cite{qin2018vins}} & \makecell[c]{VINS-RGBD \cite{shan2019rgbd}}  & \makecell[c]{VIW-Fusion }  & \makecell[c]{TartanVO \cite{wang2021tartanvo}}  & \makecell[c]{ESKF} & \makecell[c]{Ours}\\
			\hline

			\makecell[c]{Office3} & \makecell[c]{0.34}  & \makecell[c]{0.31}  & \makecell[c]{0.18 }& \makecell[c]{{1.52}} & \makecell[c]{{0.15}}  & \makecell[c]{\textbf{0.14} }  \\

		\makecell[c]{Darkroom2} & \makecell[c]{86.06 }  & \makecell[c]{0.82}  & \makecell[c]{0.53}  & \makecell[c]{1.18} & \makecell[c]{0.38 } & \makecell[c]{{\textbf{0.22}} } \\
			
		\makecell[c]{Wall2} & \makecell[c]{1.21 }  & \makecell[c]{1.00 }  & \makecell[c]{0.15 } & \makecell[c]{2.76 }  & \makecell[c]{0.20}  & \makecell[c]{\textbf{0.12}}\\
			
		\makecell[c]{Hall1} & \makecell[c]{7.06 }  & \makecell[c]{94.27 }  & \makecell[c]{{0.85} } & \makecell[c]{3.08 }  & \makecell[c]{1.29 } & \makecell[c]{\textbf{0.36} } \\
			
		\makecell[c]{Rotation3} & \makecell[c]{29.12 }  & \makecell[c]{0.19 }  & \makecell[c]{0.18 } & \makecell[c]{0.13 } & \makecell[c]{{0.14}}  & \makecell[c]{\textbf{0.08} }  \\

		\makecell[c]{Motionblur3} & \makecell[c]{9.37 }  & \makecell[c]{32.31 }  & \makecell[c]{0.78 } & \makecell[c]{1.61 }  & \makecell[c]{0.44 } & \makecell[c]{\textbf{0.26}}  \\

	\makecell[c]{Occlusion4} & \makecell[c]{--- }  & \makecell[c]{---}  & \makecell[c]{--- }  & \makecell[c]{2.35 } & \makecell[c]{0.16 }  & \makecell[c]{\textbf{0.15} } \\
	
			\hline

		\makecell[c]{Roughroad3} & \makecell[c]{0.17 }  & \makecell[c]{25.52 }  & \makecell[c]{0.14}& \makecell[c]{0.41 }  & \makecell[c]{{0.17} }  & \makecell[c]{\textbf{0.11} } \\

		\makecell[c]{Slope1} & \makecell[c]{9.41}  & \makecell[c]{2.84 }  & \makecell[c]{0.65 } & \makecell[c]{3.13 } & \makecell[c]{3.13 } & \makecell[c]{\textbf{0.64}}  \\

		\makecell[c]{Loop2} & \makecell[c]{6.09 }  & \makecell[c]{ 3.44 }  & \makecell[c]{9.23 } & \makecell[c]{13.18 } & \makecell[c]{6.31}  & \makecell[c]{\textbf{2.28 }}\\

		\makecell[c]{Corridor1} & \makecell[c]{4.48}  & \makecell[c]{{0.85} }  & \makecell[c]{1.12} & \makecell[c]{2.05 }  & \makecell[c]{1.55}& \makecell[c]{\textbf{0.74} }  \\

		\makecell[c]{Static1} & \makecell[c]{--- }  & \makecell[c]{---  }  & \makecell[c]{--- } & \makecell[c]{\textbf{0.01} } & \makecell[c]{2.87 }  & \makecell[c]{\textbf{0.01 }} \\
			\hline
	
		\end{tabular}
		\end{adjustbox}
		
		\vspace{-4mm}
	\end{table}
\vspace{-1.0mm}

\textbf{Visual challenges:}
We further introduce two new sequences for evaluation: Sequence $Hall1$ was recorded in a highly dynamic hall with a lot of people moving around; Sequence $Rotation3$ involves the robot rotating without much translation, which can influence the triangulation process in the visual front-end.
We evaluated cutting-edge SLAM systems along with our method on the aforementioned sequences. The evaluated algorithms include VINS-Mono \cite{qin2018vins}, VINS-RGBD \cite{shan2019rgbd}, VIW-Fusion \footnote{\href{https://github.com/TouchDeeper/VIW-Fusion}{https://github.com/TouchDeeper/VIW-Fusion}} and TartanVO\cite{wang2021tartanvo}(a learning-based vo). Additionally, we implement an ESKF-based IMU-wheel fusion odometry as a baseline without visual input.
The results are shown in Table \ref{ate rmse tab}, and trajectories for some sample sequences are visualized in Figure \ref{plots}. Overall, Ground-Fusion achieves the best localization results across all tested sequences.

\begin{figure}
	\small
		\begin{center}
			\begin{tabular}{cc}
			\includegraphics[scale=0.24]{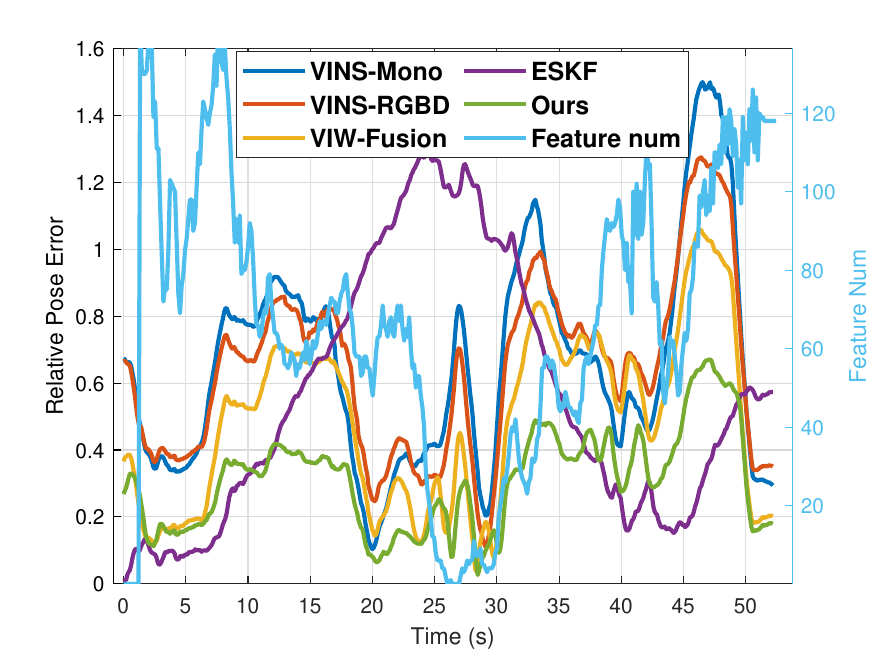} &
				\includegraphics[scale=0.24]{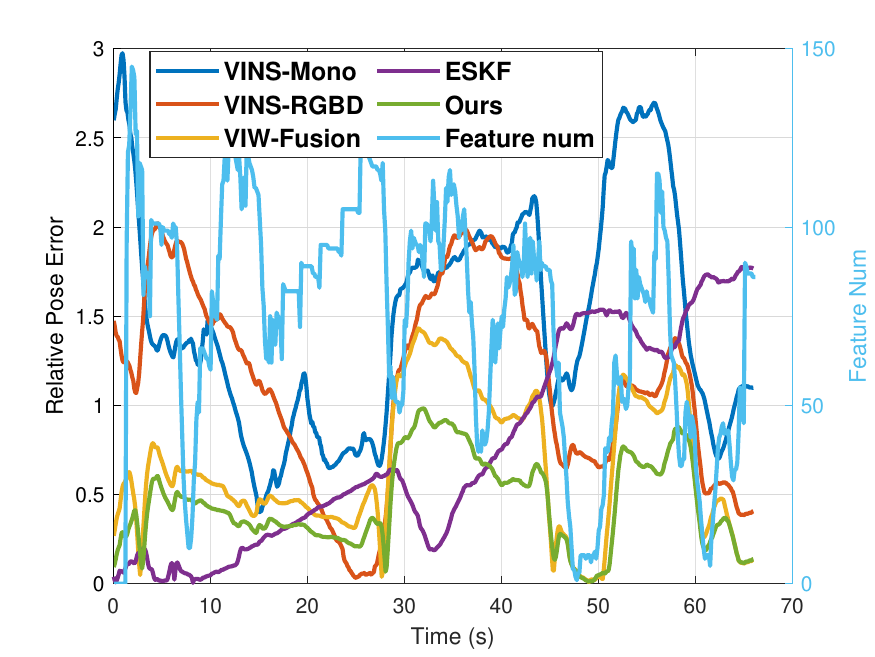} \\
				(a). Office3 & (b). Wall2\\
				
				\includegraphics[scale=0.24]{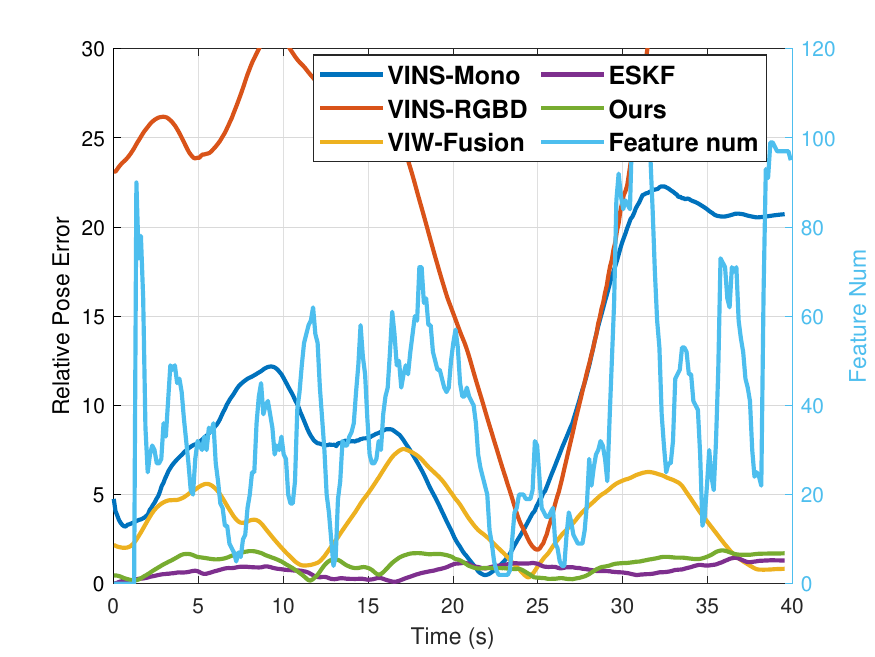}&
				\includegraphics[scale=0.24]{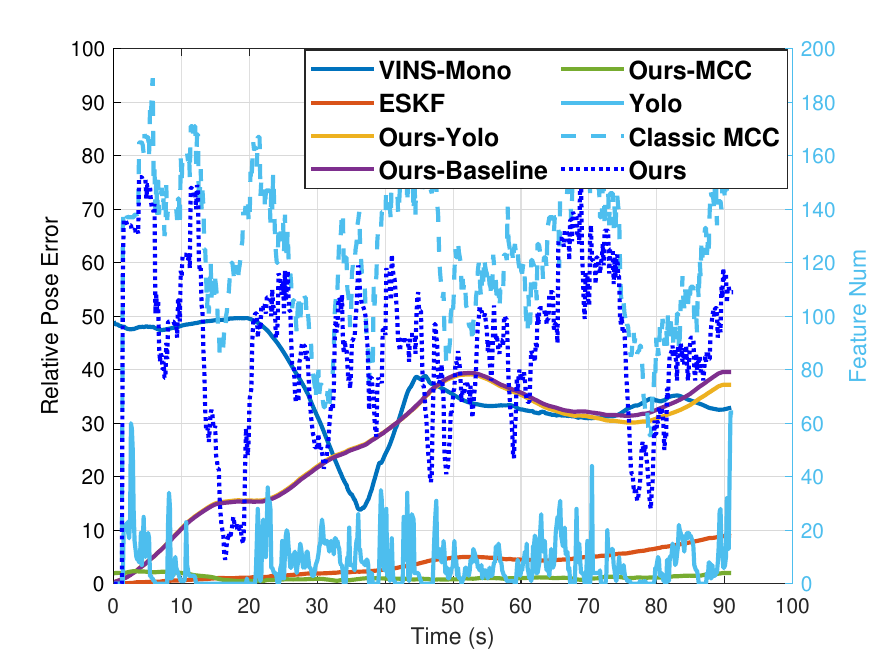} 
				\\
				(c). Motionblur3 & (d). Hall1\\

			\end{tabular}
		\end{center}
		\vspace{-3mm}
		\caption{The relative pose errors (m) of each method and  the number of effective feature points over time on some visual challenging sequences are plotted. }
		\vspace{-5mm}
		\label{point}
	\end{figure}

To illustrate the visual challenges, we plot the relative pose error (RPE) of each method and the number of valid visual feature points over time in Figure \ref{point}. The results show that insufficient feature points greatly degrade the performance of the visual front-end. For instance, at 25 seconds in Sequence $Office3$, the feature point number suddenly drops to zero due to loss, causing a notable increase in RPE for VINS-Mono \cite{qin2018vins} and VINS-RGBD \cite{shan2019rgbd}. Similarly, VINS-Mono struggles to estimate the depth by triangulation during pure rotation ($Rotation3$), resulting in significant drift. In such scenarios, our system still performs well due to the tightly-coupled wheel odometer. In $Occlusion4$ with no valid feature points observed, most systems including VIW-Fusion, fail initialization. By contrast, our system initializes using a wheel-aided dynamic approach and outperforms the wheel-IMU fusion ESKF baseline in localization accuracy.

For the dynamic environment ($Hall1$), VINS-Mono \cite{qin2018vins} suffers from significant drift. 
We conducted a comparison between the method using a YOLO \cite{redmon2018yolov3} module to remove features within the bounding box of pre-defined moving categories (e.g. people) and a baseline method with a classic MCC using the optimized pose. In Figure \ref{point} (d), while YOLO significantly reduces the number of feature points, but some of them are not actually on dynamic objects. Consequently, the RPE does not show a significant decrease when compared to the baseline. When incorporating wheel-aided MCC method, the system effectively eliminates genuine dynamic points, leading to a significant improvement in positioning accuracy. To further validate the efficacy of wheel-aided MCC, we conducted ablation tests on all seven visual challenge sequences, showing an average decrease of $\textbf{0.07}$m in the ATE RMSE compared to the baseline method.

\textbf{Wheel odometer challenges:}
In sequences $Roughroad3$ and $Slope1$, the robot moves on a rough road and a steep slope respectively; In $Loop2$, the robot traverses a hall with a carpeted aisle loop, where wheels slip considerably; Sequence $Corridor1$ is a zigzag route in a long corridor.
Table \ref{ate rmse tab} shows that our method achieves the best performance in all these sequences. 
We further conducted ablation tests to verify the effect of IMU angular velocity as a substitute for wheel angular velocity. We selected two sequences with sharp turns, including $corridor1$ and $loop2$. The results in Table \ref{angular} indicate that the IMU-odometer measurements contribute to a better localization accuracy.

	\begin{table}[h]
	\vspace{-2.5mm}
		\caption{ATE RMSE(m) of SLAM systems on selected sequences}
		\label{angular}
		\centering
		\begin{adjustbox}{width=0.63\columnwidth}
		\begin{tabular}{ccccc}
			\hline
			\makecell[c]{ Seq.} & \makecell[c]{Odom}&
			\makecell[c]{IMU-Odom}&
			\makecell[c]{Baseline} & \makecell[c]{Ours}  \\
			\hline
	
			\makecell[c]{ Corridor1} & \makecell[c]{2.17}&
			\makecell[c]{1.03}&
			\makecell[c]{0.89} & \makecell[c]{\textbf{0.66}}  \\
			
			\makecell[c]{ Loop2} & \makecell[c]{17.60}&
			\makecell[c]{6.53}&
			\makecell[c]{4.66} & \makecell[c]{\textbf{1.53}}  \\
			\hline
   			\makecell[c]{ Anomaly} & \makecell[c]{0.78}&
			\makecell[c]{0.77}&
			\makecell[c]{0.62} & \makecell[c]{\textbf{0.07}}  \\

			\makecell[c]{ Static} & \makecell[c]{ 3.54 }&
			\makecell[c]{ 3.51 }&
			\makecell[c]{  2.35 } & \makecell[c]{\textbf{0.01}}  \\
			\hline
		\end{tabular}
		\end{adjustbox}
		\vspace{-3mm}
	\end{table}

\begin{figure}
	\small
		\begin{center}
			\begin{tabular}{cc}
			\includegraphics[scale=0.24]{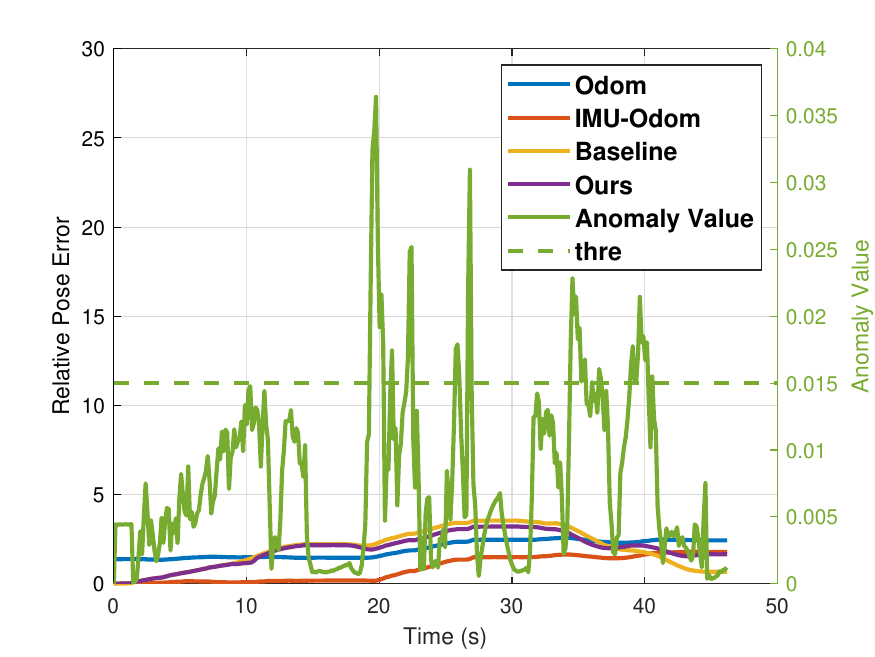} &
				\includegraphics[scale=0.24]{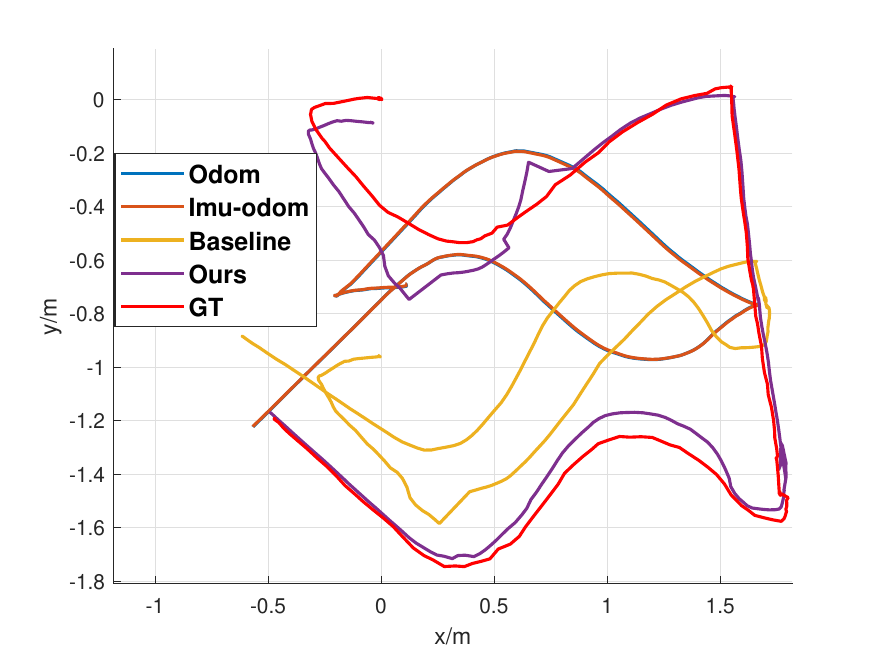} \\
				(a) & (b)\\
						
			\end{tabular}
		\end{center}
		\caption{(a) Wheel anomaly analysis and (b) Trajectory of different methods in the $Anomaly$ sequence.}
		\label{wheelc}
		\vspace{-4mm}
	\end{figure}

Moreover, we test our method on sequences with wheel anomaly. In the $Anomaly$ sequence, the robot body moves as the carpet beneath it is pulled away, while the robot wheels do not move. Conversely, in the $Static$ sequence the robot is suspended so the robot body will not move even when wheels are moving. 
The results of different methods on the two sequences are shown in Table \ref{angular}, where "baseline" denotes Ground-Fusion without wheel anomaly detection, while "ours" denotes our full method. 
As depicted in Figure \ref{wheelc} (a), a wheel anomaly is evident between 20s and 40s. Our full method adeptly eliminates erroneous wheel odometer readings here. Figure \ref{wheelc} (b) shows that only our full method matches well with the ground-truth trajectory. Similarly, our method effectively detects the wheel anomaly in the $Static$ sequence.

	\vspace{-2mm}

			




	\begin{figure}
	\small
		\begin{center}
			\begin{tabular}{cc}
			\includegraphics[scale=0.24]{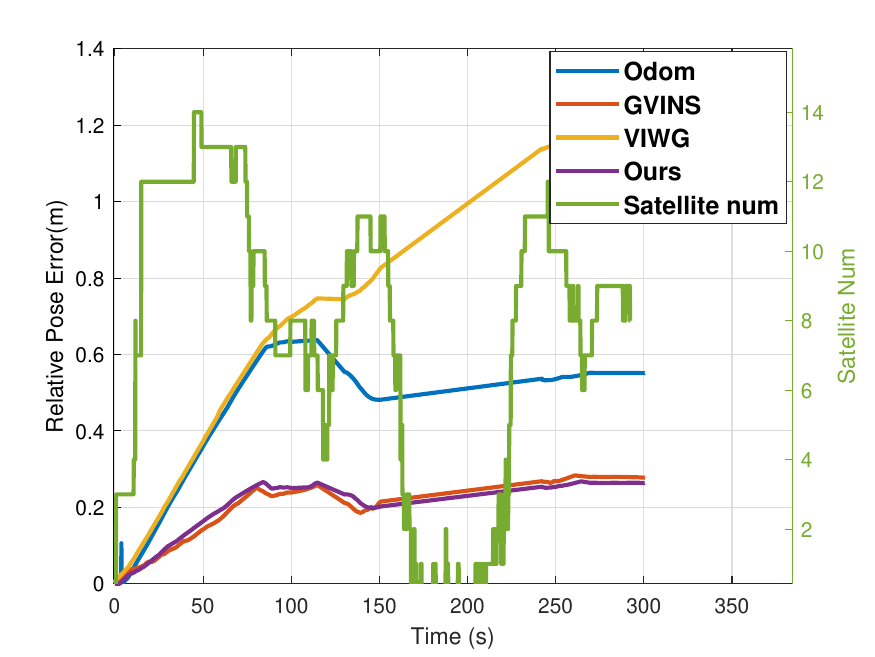} &
				\includegraphics[scale=0.17]{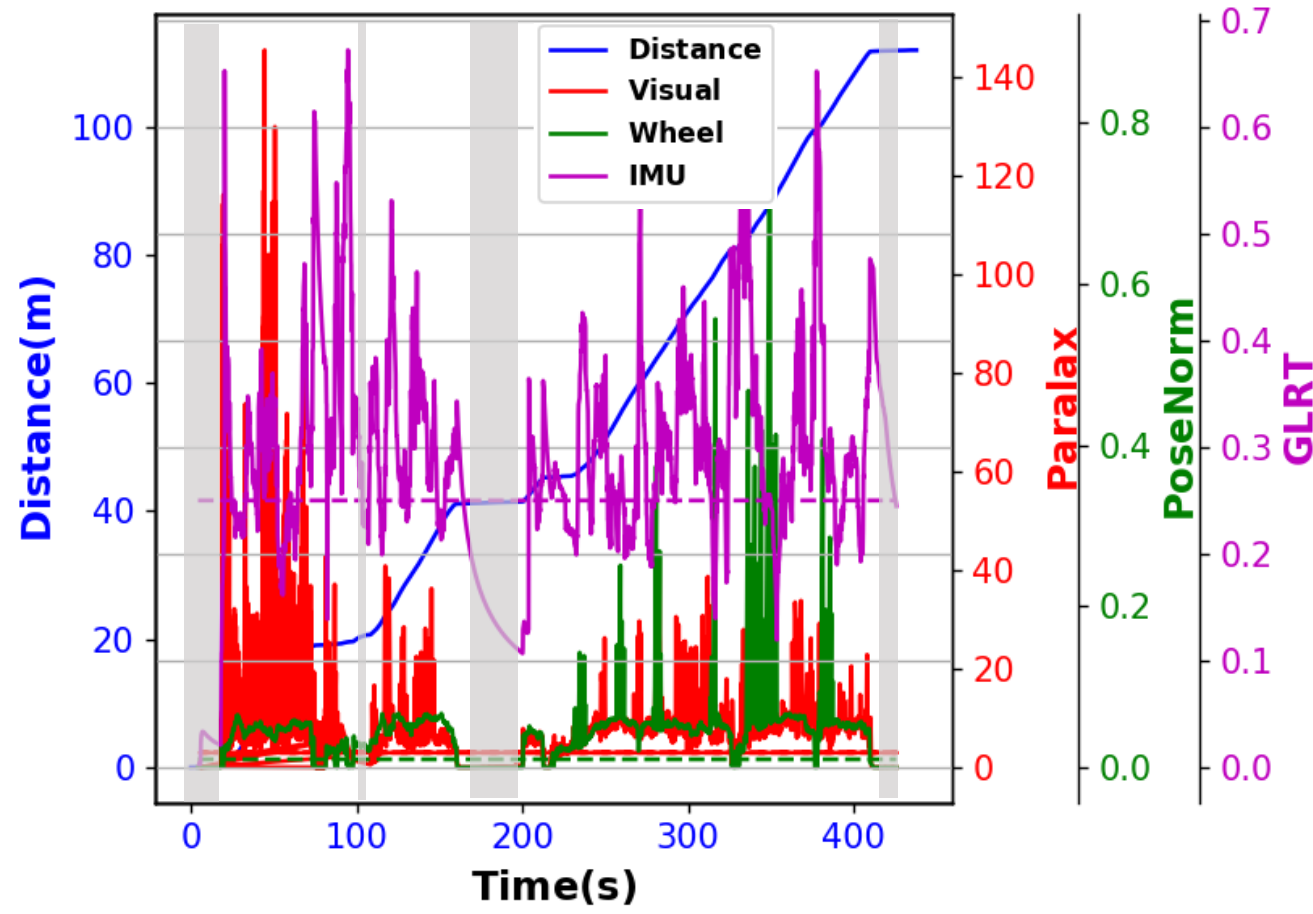} \\
				(a) & (b)\\
			\end{tabular}
		\end{center}
		\caption{(a) ATE RMSE(m) of SLAM systems on wheel anomaly sequences (b) The solid lines denote the value of each method, and dashed lines denote their corresponding thresholds. Grey shading denotes areas where at least two stationary conditions are satisfied. }
		\label{twofig}
	\end{figure}

\subsection{Outdoor Tests}

We further evaluate our method in large-scale outdoor environments as follows.
we built a ground robot for data collection,
with all the sensors well-synchronized and calibrated. We recorded some sequences in various scenarios\footnote{\href{https://github.com/SJTU-ViSyS/M2DGR-plus}{https://github.com/SJTU-ViSYS/M2DGR-plus}} and choose three most challenging sequences in this paper: In sequence $Lowspeed$, the ground vehicle moved at a low speed and made several stops; Sequence $Tree$ was under dense tree cover, causing occlusion of the GNSS satellites; In sequence $Switch$, the vehicle transitioned from outdoors to indoors, and then returned outdoors again.

\textbf{GNSS challenge:}
We evaluate our method under GNSS challenges against baseline methods, with their localization results shown in Table \ref{wt rmse tab}. Overall, Ground-Fusion outperforms baseline methods in all these cases. In $Lowspeed$, when the robot is stationary, GVINS fails to localize due to Doppler noise, and the VINS-GW \footnote{\href{https://github.com/Wallong/VINS-GPS-Wheel}{https://github.com/Wallong/VINS-GPS-Wheel}} also drifts, while Ground-Fusion works robustly by removing unreliable GNSS measurements. As shown in Figure \ref{twofig}(a), in $Switch$, VINS-GW experiences severe drift due to the loosely-coupled GNSS signals that deteriorate significantly as the robot approaches indoor areas, while both GVINS and our method remain unaffected due to their tightly-coupled integration. In Sequence $Tree$, GVINS falters due to unstable visual features, while our method remains robust due to tightly-coupled wheel and depth measurement.
	\begin{table}[h]
		\caption{ATE RMSE(m)  of SLAM systems on sample sequences}
		\label{wt rmse tab}
		\centering
		\begin{adjustbox}{width=0.64\columnwidth}
		\begin{tabular}{cccc}
			\hline
			\makecell[c]{ Sequence } & \makecell[c]{Lowspeed}& \makecell[c]{Switch} & \makecell[c]{Tree}     \\
			\hline

			\makecell[c]{Raw Odom} & \makecell[c]{8.88}  & \makecell[c]{4.95}  & \makecell[c]{{2.88}} \\

			\makecell[c]{SPP} & \makecell[c]{2.54}  & \makecell[c]{6.60}  & \makecell[c]{3.03 }  \\
			
			\makecell[c]{GVINS \cite{cao2022gvins}} & \makecell[c]{fail}  & \makecell[c]{1.40}  & \makecell[c]{{fail}} \\
			

\makecell[c]{VINS-RGBD \cite{shan2019rgbd}} & \makecell[c]{4.72}  & \makecell[c]{{1.70}} & \makecell[c]{2.27} \\
			
			\makecell[c]{VINS-GW } & \makecell[c]{20.68}  & \makecell[c]{{33.61}}& \makecell[c]{3.26}  \\


			\makecell[c]{Ours} & \makecell[c]{\textbf{0.63}}  & \makecell[c]{\textbf{1.32}}  & \makecell[c]{\textbf{0.55}} \\
			\hline

		\end{tabular}
		\end{adjustbox}
	\end{table}

\textbf{Zero velocity
update (ZUPT):}
We visualize the three stationary detection methods with the GT distance on the $Lowspeed$ sequence in Figure \ref{twofig}(b). The figure shows that a single sensor might misclassify the motion state. For instance, the wheel method fails to detect the stationary state at approximately 110 seconds. By contrast, our scheme combines three sensors, presenting reliable detection of the stationary state. Quantitatively, the ATE RMSE in $Lowspeed$ decreased by $\textbf{0.05m}$ after ZUPT.

\section{Conclusion}
This paper presents a tightly-coupled RGBD-Wheel-IMU-GNSS SLAM system to achieve reliable localization for ground vehicles. Our system features robust initialization through three strategies. We have also devised effective anomaly detection and handling methods to address corner cases, with experimental results demonstrating the superiority of our system.




\bibliographystyle{IEEEtrans}
\bibliography{root}

\end{document}